\documentclass[11pt,a4paper]{article}
\usepackage[hyperref]{eacl2021}
\usepackage{times}
\usepackage{latexsym}
\usepackage{amsmath}
\usepackage{multirow}
\usepackage{todonotes}
\usepackage{tabularx}
\usepackage{tikz}
\usepackage{array}
\usepackage{booktabs}
\usepackage{makecell}
\usepackage{siunitx}
\usepackage[basic,italic,defaultimath,nohbar,defaultmathsizes]{mathastext}
\usepackage{xcolor,colortbl}
\usepackage{hhline}

\definecolor{Gray}{gray}{0.9}
\newcolumntype{P}[1]{>{\raggedright\arraybackslash}p{#1}}
\usetikzlibrary{shapes.geometric, arrows}
\tikzstyle{arrow} = [thick,->,>=stealth]
\tikzstyle{startstop} = [rectangle, rounded corners, minimum width=2cm, minimum height=1cm,text centered, draw=black, fill=red!30, text width=2cm]

\newcolumntype{?}{!{\vrule width 1pt}}
\newcolumntype{P}[1]{>{\centering\arraybackslash}p{#1}}
\newcolumntype{M}[1]{>{\centering\arraybackslash}m{#1}}

\usepackage{microtype}

\aclfinalcopy 
\def\aclpaperid{442} 

\title{Detecting Extraneous Content in Podcasts}

\author{Sravana Reddy \\
  Spotify \\
   \texttt{sravana@spotify.com} \\\And
  Yongze Yu \\
  Spotify \\
  \texttt{yongzey@spotify.com} \\\And
   Aasish Pappu \\
  Spotify \\
  \texttt{aasishp@spotify.com} \\\AND
    Aswin Sivaraman\footnotemark[1] \\
  Indiana University \\
  \texttt{asivara@iu.edu} \\\And
 Rezvaneh Rezapour\thanks{$^{*}$\ Work done while at Spotify} \\
  University of Illinois \\at Urbana-Champaign \\
  \texttt{rezapou2@illinois.edu} \\\And
      Rosie Jones \\
  Spotify \\
  \texttt{rjones@spotify.com}
  }

\date{}

\begin{document}
\maketitle
\begin{abstract}
Podcast episodes often contain material extraneous to the main content, such as advertisements, interleaved within the audio and the written descriptions. 
We present classifiers that leverage both textual and listening patterns in order to detect such content in podcast descriptions and audio transcripts.
We demonstrate that our models are effective by evaluating them on the downstream task of podcast summarization and show that we can substantively improve ROUGE scores and reduce the extraneous content generated in the summaries.
\end{abstract}

\section{Introduction}

Podcasts are a rich source of data for speech and natural language processing. We consider two types of textual information associated with a podcast episode: the short description written by the podcast creator, and the transcript of its audio, both of which may contain content that is not directly related to the main themes of the podcasts. Such content may come in the form of sponsor advertisements, promotions of other podcasts, or mentions of the speakers' websites and products.

While such content is tightly integrated into the user experience and monetization, it is a source of noise for many natural language processing and information retrieval applications which utilize podcast data. 
For example (Table \ref{tab:motivegs}), an episode of the podcast show {\em Survival} includes a promotion for an unrelated podcast {\em Dog Tales} about dogs; a search query for podcasts on dogs should probably not surface the {\em Survival} episode. 
Algorithms attempting to connect topics discussed in the podcast to those mentioned in the episode description, such as summarization models, would be confounded by the presence of supplementary material and URLs in the description. Information extraction models looking for entities may mistakenly retrieve sponsor names from advertisements.

\begin{table}[]
\footnotesize
    \centering
    \begin{tabular}{|p{3in}|}
             \hline
         ... {\bf sit stay and roll over with excitement for par casts endearing series dog tails. Listen to dog tails free on Spotify or wherever you get your podcast.} And now back to the story. Almost immediately after setting sail on June 29th. 1871 Charles Francis Halls Arctic Expedition...\\
    \hline
      ... the focus is on strengthening deficient repertoires, while systematically increasing task demands and difficulty. {\bf For more information, visit www.behaviorbabe.com. --- This episode is sponsored by Anchor: The easiest way to make a podcast.  https://anchor.fm/app }   \\
         \hline
    \end{tabular}
    \caption{Portions of a podcast transcript and episode description which contain extraneous material (in bold).}
    \label{tab:motivegs}
\end{table}

In this paper, we introduce the problem of detecting extraneous content (which we sometimes shorthand as EC) in episode descriptions and audio transcripts. 
We produce an annotated corpus by taking advantage of podcast listening data, construct models to detect extraneous content, and evaluate our models for accuracy of detection, as well as for the downstream task of summarizing podcast transcripts. We also discuss some of challenges that arise while annotating and classifying extraneous content in this domain.

\section{Previous Work}
A related, well-studied problem is boilerplate detection on web pages, mainly involving the detection of templates, navigational elements, and advertisements \cite{kohlschutter2010boilerplate}. 
Such models tend to rely on the specific structure of web page boilerplate markup and characteristics.
There has also been work on detecting promotional content on Wikipedia \cite{bhosale-etal-2013-detecting}.

There are primarily two lines of work in advertisement detection and discovery in multimedia. One computes acoustic features over the entire audio to discriminate between the segments of content and the segments of advertisements \cite{conejero2008tv,melamed2009automatic,nguyen2010efficient}.
The other fuses multimodal features such as visual cues to segment ad clips from televised and online videos \cite{lienhart1997detection,duan2006segmentation,vedula2017multimodal}. Our work is closely related to \newcite{huang2018measuring}, who analyze consumer engagement on audio advertisements when compared to topical content, as we utilize the user engagement signals to predict extraneous content segments.

\section{Datasets and Annotation}
\label{sec:corpus}

To create an annotated dataset, we selected a random set of podcast episodes out of the Spotify Podcast Dataset, a corpus of \num{105360} episodes \cite{clifton-etal-2020-100000,trec2020podcastnotebook}. Each episode in the dataset has an automatically generated transcript and a short text description of the episode written by the podcast creators. We annotate both sources, creating ground-truth labels for the extraneous content detection task, using the open source software \textit{doccano}~\cite{doccano}. Annotators were instructed to  select spans that correspond to extraneous content, which we defined as ads, social media links, promotions of other podcasts, and show notes that are not directly related to the episode. Respecting sentence boundaries was encouraged, but not required, to allow for cases where extraneous content starts or ends mid-sentence.

\subsection{Podcast Episode Descriptions}
Annotation of episode descriptions was relatively straightforward. We encountered a few corner cases: for example, an episode may in its entirety be a promotion or an ad; a description that reflects these promotions would be on-topic. In such scenarios, the annotations attempted to be consistent with the annotators' judgments of the main topics of the episode as far as possible.
Examples of annotated episode descriptions are shown in Table \ref{tab:descegs} in the appendix.

\subsection{Podcast Episode Transcripts}
\label{sec:dips}
\begin{figure}[h]
\footnotesize
    \centering
    \includegraphics[width=3in]{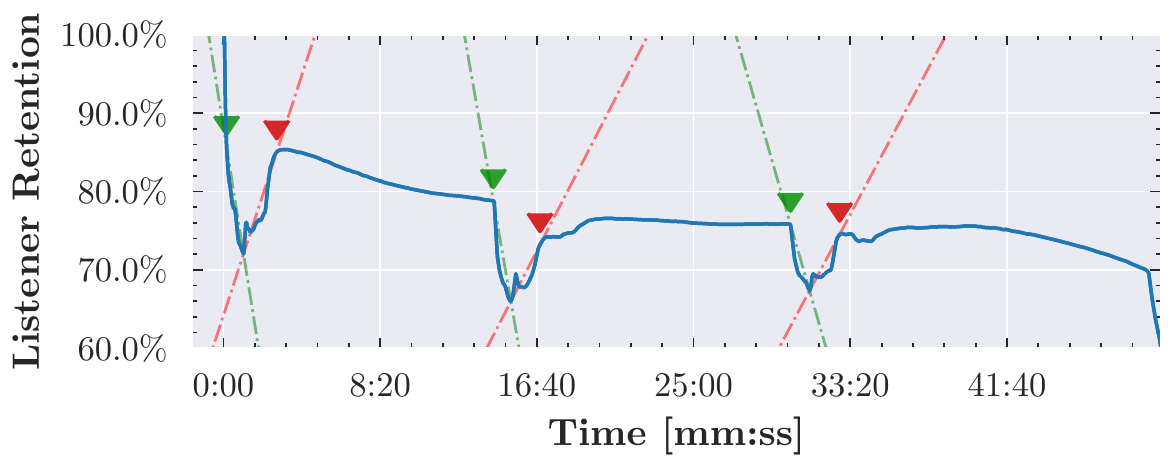}
    \caption{A podcast listener retention curve for a single podcast episode. The dips in the graph suggest potential EC regions. Start- and end-points for each dip are automatically estimated as described in Section \ref{sec:dips}, shown with green and red markers respectively.}
    \label{fig:retention}
\end{figure}

Each transcript contains thousands of words, of which extraneous content may make up a small portion. This necessitates a way to sample an informative subset of transcript segments for annotation. We observe that if a region is extraneous to the main content, listeners of the podcast episode may fast-forward past this region or abandon listening.
To this end, we gather listener data for a subset of our corpus from Spotify, an audio streaming platform. We record the proportion of all listeners who begin streaming an episode retained at each second of the episode's duration. Listening data was aggregated from the date of each episode's publication -- with the most recent episode published in February 2020 -- through September 2020. For episodes with a substantial number of listeners, there exist distinct local minima or ``dips'' in the retention curves (Figure \ref{fig:retention}), which we posit may correspond to regions of extraneous content.

To locate the center point of each dip, we first apply \texttt{SciPy} peak detection on the negative retention curve \cite{2020SciPy-NMeth}. Within \SI{\pm 2}{\min} of each peak, we calculate the slopes of secant lines passing through every point on the curve. The coordinates which maximize the secant slope within this range correspond to the start/end points of the dips. Green and red secant lines are shown in Figure \ref{fig:retention} to illustrate this process.

The transcript is then segmented \SI{60}{\sec} before the starting point and \SI{90}{\sec} after the ending point of each dip, loosening the boundaries of the potential EC regions; these segments are then manually annotated. We note that the transcripts contain noisy text which are artifacts from an automatic speech recognition (ASR) system. An additional challenge stems from identifying native advertising, where podcast creators deliberately script product placement into the context of their content \cite{einstein2015native, digiday}. Considering all cases, we attempted to best estimate the boundaries of the scripted content. 

Examples of annotated transcript regions are shown in Table \ref{tab:contentegs} in the appendix.
The set of manual annotations is used as a {\em gold} labeled transcript dataset for our models.

Of the annotated dips, \SI{38.4}{\percent} were found not to contain extraneous content. These largely correspond to episode beginnings, where listeners may skip over the introduction to show, or dynamically inserted ads that are not present in the transcript. The dip boundaries for the rest are relatively accurate against the manual annotations, with an absolute mean error of \num{16.0} words for the starts, and \num{35.2} for the ends, motivating the use of the un-annotated listener dips as a {\em silver} training set (described in \S\ref{sec:detectors:trans}).

\paragraph{Sentence-Level Labels}
Our models for descriptions and transcripts use the sentence as the unit of classification.
The annotated data is split into sentences using SpaCy\footnote{https://spacy.io}. A sentence is labeled as extraneous if more than \SI{50}{\percent} is annotated as such (Table \ref{tab:annotationcounts}).

\begin{table}[]
\centering
\resizebox{\linewidth}{!}{
\renewcommand{\arraystretch}{1.5}
\begin{tabular}{|M{2.5cm}||r|r|}
\Xhline{4\arrayrulewidth} 
\textbf{Dataset} &
\textbf{\# of episodes} &
\textbf{\# of sentences} \\
\Xhline{4\arrayrulewidth}
\textit{\shortstack{\\Episode\\Descriptions}}
& 1730 \small{(61.58\% w/ EC)} & 10410 \small{(43.80\% w/ EC)} \\
\hline
\textit{\shortstack{\\Transcripts\\(Gold)}}
& 506 \small{(62.25\% w/ EC)} & 12918 \small{(24.01\% w/ EC)} \\
\Xhline{4\arrayrulewidth} 
\end{tabular}}
\renewcommand{\arraystretch}{1}
\caption{Our manually annotated datasets show that the majority of podcast episodes do contain extraneous content, making up a sizeable fraction of all sentences.}
\label{tab:annotationcounts}
\end{table}

\section{Extraneous Content Detection}
\label{sec:detectors}

\subsection{Sentence-Level Classification}
We built separate classifiers for detecting extraneous content in descriptions and transcripts. A pre-trained BERT \cite{devlin-etal-2019-bert} cased language model\footnote{https://huggingface.co/bert-base-cased} was first fine-tuned on our entire large corpus (of podcast descriptions and transcripts respectively, excluding the test set) to capture the distinctive language use of the podcast domain, and then further fine-tuned for classification on the annotated data to predict whether a sentence is extraneous.
We also trained non-neural classifiers (logistic regression and SVMs) with TF-IDF unigram and bigram features (Appendix \ref{sec:modeldetails}).

We experimented with single sentence classification in isolation, and with the immediately preceding sentence prepended for context.\footnote{The two sentences are concatenated. In BERT, we include the SEP token to separate them. Where there is no preceding sentence, we use a special {\tt \_\_START\_\_} token.}

\paragraph{Results}
Our experiments used a training/test split of \num{90}/\num{10} for annotated transcripts and \num{80}/\num{20} for episode descriptions. The best results come from BERT where the model sees the previous sentence (Table \ref{tab:sentmetric}). We observed that classification precision and recall are comparable on the descriptions dataset; however, with transcripts, precision tends to be lower (\num{0.690}) than recall (\num{0.870}). This suggests that the model mistakenly identifies sentences as extraneous, but has fairly good coverage. This seems to be because many sentences in extraneous content in the training transcripts are seemingly innocuous when taken out of context, confounding the model during training.

\begin{table}[t]
  \centering
    \resizebox{\linewidth}{!}{
    \renewcommand{\arraystretch}{1.5}
    \begin{tabular}{|c|c||r|r||r|r|}
\Xhline{4\arrayrulewidth}
 &  & \multicolumn{2}{c||}{\textbf{Single Sentence}} & \multicolumn{2}{c|}{\textbf{Sentence with Context}} \\ \cline{3-6}
{\multirow{-2}{*}{\textbf{Dataset}}} & {\multirow{-2}{*}{\textbf{Model}}} & \textbf{Unigram} & \textbf{Bigram} & \textbf{Unigram} & \textbf{Bigram} \\

\Xhline{4\arrayrulewidth}

{\multirow{3}{*}{\textit{\shortstack{Episode\\Descriptions}}}} & LR  & 0.893 & 0.818 & 0.858 & 0.850 \\ \cline{2-6}
                                                               & SVM & 0.907 & 0.907 & 0.898 & 0.896 \\ \cline{2-6}
                                                               & BERT & \multicolumn{2}{c||}{0.920} & \multicolumn{2}{c|}{\textbf{0.940}} \\
\Xhline{4\arrayrulewidth}
{\multirow{3}{*}{\textit{\shortstack{Transcripts\\(Gold)}}}} & LR  & 0.651 & 0.603 & 0.629 & 0.638 \\ \cline{2-6}
                                                             & SVM & 0.675 & 0.679 & 0.663 & 0.670 \\ \cline{2-6}
                                                             & BERT & \multicolumn{2}{c||}{0.710} & \multicolumn{2}{c|}{\textbf{0.769}} \\
\Xhline{4\arrayrulewidth}
\end{tabular}}
    \renewcommand{\arraystretch}{1}
\caption{F1 score of three sentence-level EC classifiers tested across two datasets.}
\label{tab:sentmetric}
\end{table}

\begin{table}[hbt]
  \centering
    \resizebox{\linewidth}{!}{
    \renewcommand{\arraystretch}{1.5}
    \begin{tabular}{|c|M{4cm}||r|r|}
\Xhline{4\arrayrulewidth}
\textbf{Dataset} & \textbf{Model} & \textbf{Doc. (\%)} & \textbf{Sen. (\%)} \\
\Xhline{4\arrayrulewidth}
{\multirow{3}{*}{\textit{\shortstack{Episode\\Descriptions}}}} & Sentence-level BERT & 86.4 & \cellcolor{white}{94.9} \\ \cline{2-4}
                                                & BERT + Change Point & \textbf{89.6} & 95.4 \\ \cline{2-4}
                                                & BERT + BiLSTM-CRF & 87.6 & 95.0 \\
\Xhline{4\arrayrulewidth}
{\multirow{2}{*}{\textit{\shortstack{Transcripts\\(Gold)}}}} & Sentence-level BERT & 43.1 & \cellcolor{white}{88.7} \\ \cline{2-4}
                                                & BERT + Smoothing & \textbf{49.1} & 90.9 \\
\Xhline{4\arrayrulewidth}
{\multirow{2}{*}{\textit{\shortstack{Transcripts\\(Silver)}}}} & Sentence-level BERT & 60.8 & 96.6 \\ \cline{2-4}
                                                & BERT + Smoothing & \textbf{66.7} & 97.0 \\
\Xhline{4\arrayrulewidth}
\end{tabular}}
    \renewcommand{\arraystretch}{1}
\caption{Classification accuracy of sentence-level and document-level extraneous content detection. Documents come from either the dataset of full episode descriptions, or from transcript segments corresponding to dip regions. A document EC-match occurs when {\em all} sentence labels agree; this is difficult to achieve with longer segments.}
\label{tab:docacc}
\end{table}

\subsection{Document-Level Classification} 

We classify sentences independently, but extraneous content comes as contiguous groups of sentences. We apply non-parametric kernel regression to post-hoc smooth the sequence of individual sentence classification probabilities in the transcripts. We observe that extraneous content within episode descriptions often appears as a contiguous block at the end, prompting us to apply a change point detector (Appendix \ref{sec:changepoint}) on the sentence classification probabilities in order to detect the start of the EC block.

We also formulate the problem as sequence tagging at the sentence level, in order to allow the model to learn the label dependencies. For this, we use the BERT pooled sentence embeddings as input to a separate BiLSTM-CRF model. The BiLSTM-CRF improves over sentence-level classification but underperforms the change point detection strategy (Table \ref{tab:docacc}).
In future work, we would like to investigate an end-to-end BERT-BiLSTM-CRF model \cite{dai2019named} or sequential sentence classification models \cite{cohan-etal-2019-pretrained}.

\paragraph{Expanded Transcript Dataset}
\label{sec:detectors:trans}

Since the manually labeled gold set is small, we create a larger silver dataset from \num{6401} detected listener dips across \num{4930} episodes by applying the best performing classifier trained on the gold data. To encode information about the dip locations to aid the model, we prepend special tokens `in-dip'
and `outside-dip' to each sentence depending on whether the sentence lies within a listening dip.  

We strip special tokens from the resulting silver set. This data is then used to train a final classifier that can detect extraneous content regions in podcast transcripts {\em without} listener dips. We also add a negative sample of sentences that are distant (by at least 5 minutes) from dips in the same episodes, with the assumption that these are
likely to be topical.

On the same test set as the previous experiments, the document and sentence level performance increases (Table \ref{tab:docacc}), proving the model benefits from the larger, albeit noisy, training set.
This model is applied to the corpus in the downstream task described below.

\section{Application to Podcast Summarization}

We address the problem of automatically generating episode descriptions from podcast transcripts, a task similar to abstractive summarization. Within this problem, we evaluate the downstream effect of removing extraneous sentences from the training and/or test data. Alternatives to removal (such as using the model's predictions as auxiliary inputs in the downstream system) are left for future work.

We experiment with two supervised abstractive summarization models both built using BART \cite{lewis-etal-2020-bart}.
The first experiment uses a model pretrained for summarization on the CNN/DailyMail dataset\footnote{https://huggingface.co/facebook/bart-large-cnn} \cite{NIPS2015_5945}. This model (which we refer to as \textsc{BART-CNN}) evaluates the extent to which extraneous text in the transcripts contribute to the presence of extraneous content in the generated descriptions.
In the second experiment, we fine-tune \textsc{BART-CNN} on our corpus of podcasts, similar to the work of \citet{zheng2020baseline}, using the episode transcripts as inputs and descriptions as targets.  
We refer to this model as \textsc{BART-Podcasts}; with it, we can evaluate the effects of EC as realistic noise which may contaminate the training data of summarization models.

\subsection{Experimental Setup}
From the original corpus of \num{105360} podcast shows, \num{6401} were used for training and evaluation of the two EC detection models. We filter the remaining episodes for descriptions which are distinctly short or long. The resulting dataset contains \num{84451} episodes sorted by episode publication date. This is split into \num{82451} episodes for training, \num{1000} for validation, and \num{1000} for evaluation. 

As a baseline, we manually remove extraneous content from the episode descriptions within the test set, comparing the {\sc Rouge} score of the model outputs against the manually-cleaned descriptions as well as against the original descriptions. Additionally, we manually validate whether the generated outputs for \num{150} random test episodes contain extraneous content.

\begin{table*}[!t]
\footnotesize
\centering
\renewcommand{\arraystretch}{1.5}
\setlength\tabcolsep{3pt}

\begin{tabular}{|M{3.6cm}|c||r|r|r||r|r|r||r|}
\Xhline{4\arrayrulewidth} 
{\multirow{2}{*}{\textbf{Model}}} &
{\multirow{2}{*}{\textbf{\shortstack{EC Removal\\ Method}}}} &
\multicolumn{3}{c||}{\shortstack{\\\textsc{Rouge-L} against\\ original descriptions}} &
\multicolumn{3}{c||}{\shortstack{\\\textsc{Rouge-L} against\\ cleaned descriptions}} &
\multirow{2}{*}{\shortstack[r]{Amount of EC \\ in output (\%)}} \\
\cline{3-8}
& & Rec. & Pre. & F & Rec. & Pre. & F & \\
\Xhline{4\arrayrulewidth} 
\textit{Original Descriptions\newline\small{(no model)}} 
& n/a & 1.000 & 1.000 & 1.000 & 0.996 & 0.810 & 0.872 & 50.0 \\
\Xhline{4\arrayrulewidth}
{\multirow{3}{*}{\textit{\shortstack{BART-CNN}}}}
& n/a & 0.310 & 0.199 & 0.204 & 0.340 & 0.170 & 0.195 & 18.8 \\ \cline{2-9}
& M1  & 0.299 & 0.200 & 0.199 & 0.330 & 0.172 & 0.192 & 11.5 \\ \cline{2-9}
& M2  & 0.309 & 0.188 & 0.195 & 0.345 & 0.165 & 0.191 & 4.0  \\
\Xhline{4\arrayrulewidth}
{\multirow{4}{*}{\textit{\shortstack{BART-Podcasts}}}}
& n/a & 0.377 & 0.329 & 0.315 & 0.349 & 0.228 & 0.248 & 73.2 \\ \cline{2-9}
& M1  & 0.269 & 0.309 & 0.247 & 0.308 & 0.280 & 0.260 & 8.7  \\ \cline{2-9}
& M3  & 0.323 & 0.276 & 0.261 & 0.367 & 0.247 & 0.263 & 15.5 \\ \cline{2-9}
& M4  & 0.330 & 0.281 & 0.269 & 0.376 & 0.261 & 0.268 & 2.0  \\
\Xhline{4\arrayrulewidth} 

\end{tabular}
\caption{Evaluation of the effect of extraneous content detection on a downstream summarization task. Accuracies are reported in terms of recall, precision, and F1 score. The models are provided with data pre-processed through one of four EC removal strategies: (M1) cleaned test descriptions, (M2) cleaned test transcripts, (M3) cleaned training descriptions, or (M4) cleaned descriptions and transcripts for both train and test.}
\label{tab:summ}
\end{table*}

\subsection{Results}
Table \ref{tab:summ} shows the full {\sc Rouge-L} scores of our experiments. We evaluate quality through {\sc Rouge} as well as by manually verifying for the presence of extraneous content in the outputs. {\sc BART-CNN} is an out of the box summarization model, while {\sc BART-Podcasts} is the same model  that is fine-tuned on our data of transcripts and descriptions. All numbers are reported on the test split of the corpus. The range of these {\sc Rouge} scores is comparable to previous podcast summarization work. Removing extraneous content is clearly beneficial for summary quality: while the baseline models have better {\sc Rouge} scores against the original (EC-containing) descriptions, the highest scoring models score better against the clean descriptions compared to the originals. 
With the original data, both {\sc BART-CNN} and {\sc BART-Podcasts} tend to generate descriptions that contain extraneous material. Interestingly, {\sc BART-Podcasts}, being trained on the unmodified descriptions, produces even {\em more} extraneous content (\SI{73.2}{\percent}) than the corresponding original descriptions (\SI{50.0}{\percent}), often generating ads unrelated to the actual sponsors, and nonexistent URLs. While post-processing only the output summaries with no change to the model inputs is effective at minimizing extraneous content, it does so at the expense of summary quality, since the resulting summaries are significantly shortened. The best overall performance comes from detecting extraneous content on transcripts and descriptions before model training and application.

\section{Conclusion}
We introduced the problem of detecting extraneous content in podcast descriptions and transcripts, presented models that leverage textual and listener data, and evaluated them on a downstream summarization task. 

We consider our models to be baselines for a new problem with several opportunities for future work.
Although we used two separate models for the descriptions and transcripts with the view that the language patterns are different, a joint model or shared components may be able to take advantage of some of the common vocabulary. One could leverage the `boilerplate' nature of some types of extraneous content like ads by detecting repeated sentences and phrases across the corpus.
A language model that is robust to noisy speech transcripts \cite{9003890,chuang2019speechbert} may improve accuracy on podcast transcripts. 
Given that extraneous content may appear as pre-recorded audio, or with a different speaking pitch and cadence, acoustic features alongside textual ones may be helpful. 

\section{Acknowledgments}
We thank Ann Clifton and Jussi Karlgren for valuable  assistance with the summarization model and evaluations, 
Richard Mitic, Laurence Pascall, Skylar Brannon, Kristie Savage, and Karishma Gulati for collaborating on identifying listening dips from user logs, Iftekhar Tanveer for helpful discussions, and the anonymous EACL reviewers for their suggestions.

\bibliography{boilerplate_refs}
\bibliographystyle{acl_natbib}

\appendix
\section{Appendix}
\subsection{Annotation Process and Annotated Examples}
\label{sec:annotationexamples}

Examples of extraneous content regions of episode descriptions are shown in Table \ref{tab:descegs}.
and example of annotated listener `dip' regions in the podcast transcripts in Table \ref{tab:contentegs}. 
The categorization into types is only for illustrative purposes and is not used in our model.

\begin{table*}[]
    \renewcommand{\arraystretch}{1.2}
    \centering
    \begin{tabular}{|P{0.8in}|P{5.0in}|}
         \hline
         \textsc{Type} & \textsc{Examples}  \\
         \hline
         Platform Ads & Transformation Church Pastor Mike Todd brings Marked Part 8- You Are Enough {\bf {\tt ---}   This episode is sponsored by Anchor: The easiest way to make a podcast.  https://anchor.fm/app  Support this podcast: https://anchor.fm/m-e8/support} \\
         \hline
         Social media links, podcast promotions, sponsor ads & Crazy new years news, Dallin fantasizes about a Tesla, and then we do some hilarious Mad Libs... you have to hear these. {\bf Follow the Dashleys for more! https://thedashleysxxxx.com https://facebook.com/thedashleysxxx https://instagram.com/thedashxxxx https://youtube.com/thedasxxxx hellodasxxxx@gmail.com Check out our new Podcast! http://anchor.fm/takingsxxxx Thank you to Skillshare for sponsoring this podcast! Get 2 months premium membership free here:  skillshare.com/dashxxxx Thank you to HelloFresh for sponsoring this podcast! Sign up for HelloFresh and get 10 free meals here!   URL: hellofresh.com/BIGLITTLELIxxx Promo: BIGLITTLELxxxx   ---   Support this podcast: https://anchor.fm/biglittlexxxx/support}\\\hline
    \end{tabular}
    \caption{Annotated examples from episode descriptions. (Some of the links are masked for privacy.)}
    \label{tab:descegs}
\end{table*}

\begin{table*}[]
    \renewcommand{\arraystretch}{1.2}
    \centering
    \begin{tabular}{|p{0.8in}|p{0.8in}|p{4.0in}|}
    \hline
    \textsc{Type} & \textsc{Subtype} & \textsc{Examples}  \\
    \hline
    Sponsorships\newline/ Ads &  Products & ... the railway tracks a mere four inches a small but deadly act of sabotage. {\bf Hey friend, I want to tell you about the new \$3 Little John from Jimmy John's [...] or with the Jimmy John's app at participating locations taxes and delivery fees extra.} Welcome to today and True Crime a par cast original...  \\
    \cline{2-3} 
    & Platforms & ... before we move on to your next topic. {\bf We just want to say thanks to our sponsor Anchor. If you haven't heard about Anchor [...] and it's everything you need to make a podcast in one place download the free anchor app or go to Anchor.} Um to get started, a very somber... \\
    \hline
    Podcast\newline Promotions & Other\newline Podcasts & ... Coming up heading returns to Denmark and Begins the long journey to reclaim his throne. {\bf Darkness tragedy pain these things hide within every beloved institution and most people are none the wiser every week. [...] Search The Dark Side of or visit Park ask.com / Dark Side to listen now.} Now back to the story...  \\
    \cline{2-3}
    & Social\newline Media & ... manipulates our relationships into deadly results {\bf at par cast we are grateful for you our listeners you allow us to do what we love let us know how we're doing reach out on Facebook and Instagram at par cast and Twitter [...] for more information.} It all began with a computer...  \\
    \hline
    Not EC & Episode\newline Introduction & ... I'm your host Taylor structure of the live Daily Talk Radio Show [...] It's getting to meet interesting inspiring and exceptional people and getting them to talk about the heaviest things in the world. So if you love real talk with a...   \\
    \cline{2-3}
    & Miscellaneous\newline Content & ... Not much is known about her personal life prior to her murder if she was so average, why was she living with her aunt and uncle I'm not saying there's anything wrong with that [...] I'm sorry to say that it looks as though your husband has an advanced... \\
    \hline
    \end{tabular}
    \caption{Annotated examples from the podcast transcripts, corresponding to detected dips in listening.}
    \label{tab:contentegs}
\end{table*}

\subsection{Model Training and Evaluation Details}
\label{sec:modeldetails}

We modified the sentence splitter in SpaCy to include {\tt ---}, {\tt ...} and the three-space string as delimiters for episode descriptions, based on our observations of common patterns. Speech recognition errors/disfluencies and missing punctuation contribute to a small amount of noise in the sentence segmentations of transcripts and descriptions respectively.

For the bag of words models, we used \url{https://scikit-learn.org/} implementations with the default parameters and no hyper-parameter tuning.

For all Transformer models, we used the Huggingface library \cite{wolf-etal-2020-transformers}. For summarization, we set the maximum length of the target descriptions as 250 tokens for training and generation, and the minimum length to 30 tokens. The models were trained for up to 5 epochs, with early stopping based on {\sc ROUGE-2} on the validation set. All other hyperparameters were set as the defaults specified in the Huggingface Transformers code.

For evaluation, we use the \textsc{Files2Rouge} implementation (\url{ https://github.com/pltrdy/files2rouge}), after tokenizing both the reference and the outputs using the Moses tokenizer (\url{ https://github.com/alvations/sacremoses}).

\subsection{Change Point Detection for Episode Descriptions}
\label{sec:changepoint}
As shown in Eq. \ref{eqn:changepoint}, we can find a position $\hat{\tau}$ which maximizes the negative log-likelihood ratio $\textit{R}_{\tau}$ of $\textit{H}_{1}$ as the existence of the change point versus $\textit{H}_{0}$ as no change point.

\begin{equation} \label{eqn:changepoint}
\hat{\tau} = \displaystyle\arg \max_{\substack{1\leq \tau \leq N}} \textit{R}_{\tau}
\end{equation}

We make the assumption that $(1)$ there is only one change point, and $(2)$ extraneous content appears at end the descriptions.
The null hypothesis is that there is no changepoint, while the alternative hypothesis assumes that there is a changepoint at the time $t=\tau$. Here is the hypothesis test:
\begin{equation} \label{eq1.1}
H_0:\theta_1 = \theta_2 = ... = \theta_{N-1} = \theta_{N} \\
\end{equation}
\begin{equation} \label{eq1.2}
H_1:\theta_1 = ... = \theta_{\tau-1} = \theta_{\tau} \neq \theta_{\tau+1}=...= \theta_{N}
\end{equation}
Therefore, the likelihood is given by the probability of observation the data $x$ = $x_1$,...,$x_n$ conditional on $H_0$. In other words, 
\begin{equation} \label{eq1.3}
\emph{L}(\textit{H}_{0}) = p(x | \textit{H}_0) = \prod_{i=1}^{N} p(x_i|\theta_i)
\end{equation}
And the the likelihood of the alternative hypothesis is,
\begin{equation} \label{eq1.4}
\emph{L}(\textit{H}_{0}) = p(x | \textit{H}_1) = \prod_{i=1}^{\tau} p(x_i|\theta_i)\prod_{j=\tau+1}^{N} p(x_j|\theta_j)
\end{equation}

The log-likelihood ratio ${R_\tau}$ is then,

\begin{equation} \label{eq1.5}\begin{split}
\textit{R}_{\tau} &= \log{\frac{\emph{L}_{\textit{H}_0}}{\emph{L}_{\textit{H}_1}}} \\
 &= \sum_{i=1}^{\tau} \log{p(x_i|\theta_1)} + \sum_{j=\tau+1}^{N}\log{p(x_j|\theta_2)} \\&\quad - \sum_{k=1}^{N}\log{p(x_k|\theta_0)}
\end{split}\end{equation}

\end{document}